\documentclass{article}

\usepackage{multicol}
\usepackage{amsthm}
\usepackage{amsmath}
\usepackage{amssymb}
\usepackage{bm}
\usepackage{mathrsfs}
\usepackage[dvips]{graphicx}

\usepackage{algorithm}
\usepackage{algorithmic}

\usepackage{color}

\usepackage{url}

\usepackage{supertabular}

\newtheorem{theorem}{Theorem}[section]

\newtheorem{proposition}[theorem]{Proposition}

\theoremstyle{definition}

\theoremstyle{remark}

\DeclareMathAlphabet{\mathsfsl}{OT1}{cmss}{m}{sl}

\renewcommand{\phi}{\varphi}

\newcommand{\argmin}{\operatorname*{arg\,min}}

\newcommand{\bW}{\boldsymbol{W}}

\renewcommand{\algorithmicrequire}{\textbf{Input:}}
\renewcommand{\algorithmicensure}{\textbf{Output:}}

\def\b0{\mathbf{0}}

\def\bR{\boldsymbol{R}}

\def\bA{\boldsymbol{A}}

\usepackage{times}
\usepackage[moderate]{savetrees}
\usepackage{amsmath}
\usepackage{amssymb}
\usepackage{microtype}
\usepackage{graphicx}
\usepackage{subfigure}
\usepackage{booktabs} 

\usepackage{hyperref}

\usepackage[accepted]{icml2020}

\icmltitlerunning{Message Passing Least Squares Framework and its Application to Rotation Synchronization}

\begin{document}

\twocolumn[
\icmltitle{Message Passing Least Squares Framework \\and its Application to Rotation Synchronization
           }



\icmlsetsymbol{equal}{*}

\begin{icmlauthorlist}
\icmlauthor{Yunpeng Shi}{umn}
\icmlauthor{Gilad Lerman}{umn}
\end{icmlauthorlist}

\icmlaffiliation{umn}{School of Mathematics, University of Minnesota, Minneapolis, MN, USA}
\icmlcorrespondingauthor{Yunpeng Shi}{shixx517@umn.edu}
\icmlcorrespondingauthor{Gilad Lerman}{lerman@umn.edu}
\icmlkeywords{Machine Learning, ICML}

\vskip 0.3in
]



    \printAffiliationsAndNotice{}  

\begin{abstract}

We propose an efficient algorithm for solving 
group synchronization under high levels of corruption and noise, while we focus on rotation synchronization. We first describe our recent theoretically guaranteed message passing algorithm that estimates the corruption levels of the measured group ratios. We then propose a novel reweighted least squares method to estimate the group elements, where the weights are initialized and iteratively updated using the estimated corruption levels. We demonstrate the superior performance of our algorithm over state-of-the-art methods for rotation synchronization using both synthetic and real data.

\end{abstract}

\section{Introduction}
The problem of group synchronization is critical for various tasks in data science, including structure from motion (SfM), simultaneous localization and mapping (SLAM), Cryo-electron microscopy imaging, sensor network localization, multi-object matching and community detection. Rotation synchronization, also known as rotation averaging, is the most common group synchronization problems in 3D reconstruction.  It asks to recover camera rotations from 
measured
relative rotations 
between
pairs of cameras. Permutation synchronization, which has applications in multi-object matching, asks to obtain globally consistent matches of objects from possibly erroneous measurements of matches between some pairs of objects. The simplest example of group synchronization is $\mathbb Z_2$ synchronization, which appears in community detection. 

The general problem of group synchronization can be mathematically formulated as follows.
Assume a graph $G([n],E)$ with $n$ vertices indexed by $[n]=\{1,\ldots,n\}$, a group $\mathcal{G}$, and a set of group elements $\{g_i^*\}_{i=1}^n \subseteq \mathcal{G}$.  
The 
problem asks to recover $\{g_i^*\}_{i=1}^n$ from noisy and corrupted measurements $\{g_{ij}\}_{ij \in E}$ of the group ratios $\{g_i^*g_j^{*-1}\}_{ij \in E}$.
We note that one can only recover, or approximate, the original group elements $\{g_i^*\}_{i\in [n]}$ up to a right group action.
Indeed, for any $g_0\in \mathcal G$, $g_{ij}^*$ can also be written as $g_i^*g_0(g_j^*g_0)^{-1}$ and thus $\{g_i^*g_0\}_{i\in [n]}$ is also a solution. 
The above mentioned synchronization problems (rotation, permutation, and $\mathbb Z_2$ synchronization) correspond to the groups $SO(3)$, $S_N$, and $\mathbb Z_2$, respectively.

The most challenging issue for group synchronization is the practical scenario of highly corrupted and noisy measurements. Traditional least squares solvers often fail to produce accurate results in such a scenario. 
Moreover, some basic estimators that seem to be robust to corruption
often do not tolerate in practice high level of noise. 
We aim to propose a general method for group synchronization that may tolerate high levels and different kinds of corruption and noise. While our basic ideas are formally general, in order to carefully refine and test them, we focus on the special problem of rotation synchronization, which is also known as rotation averaging \cite{RotationAveraging13}.
We choose this problem as it is the most common, and possibly most difficult, synchronization problem in 3D computer vision.

\subsection{Related Works}
\label{sec:related}
Most previous group synchronization solvers minimize an energy function. 
For the discrete groups $\mathbb Z_2$ and $S_N$, least squares energy minimization is commonly used.
Relevant robustness results, under special corruption and noise models, are discussed in  \citet{Z2Afonso2, Z2abbe, Z2Afonso, chen_partial, Huang13, PPM_vahan, deepti}.

For Lie groups, such as $SO(D)$, that is,  the group of $D \times D$ orthogonal matrices with determinant 1, where $D\geq 2$, least squares minimization was proposed to handle Gaussian noise \citep{rotationNP,StrongDuality18,Govindu04_Lie}. However, when the measurements are also adversarially corrupted, this framework does not work well and other corruption-robust energy functions need to be used \cite{ChatterjeeG13_rotation, L12, HartleyAT11_rotation, SO2ML, wang2013exact}. 
The most common corruption-robust energy function uses least absolute deviations.
\citet{wang2013exact} prove that under a very special probabilistic setting with $\mathcal G = SO(D)$, the pure minimizer of this energy function can exactly recover the underlying group elements with high probability. However, their assumptions are strong and they use convex relaxation, which changes the original problem and is expensive to compute. 
 \citet{SO2ML} apply a trimmed averaging procedure for robustly solving $SO(2)$ synchronization. They are able to recover the ground truth group elements under a special deterministic condition on the topology of the corrupted subgraph. However, the verification of this condition and its extension to $SO(D)$, where $D>2$, are nontrivial.
\citet{HartleyAT11_rotation} used the Weiszfeld algorithm
 to minimize the least-absolute-deviations energy function
with $\mathcal G = SO(3)$. Their method iteratively computes geodesic medians. However, they update only one rotation matrix per iteration, which results in slow empirical convergence and may increase the possibility of getting stuck at local minima.  \citet{L12} proposed a robust Lie-algebraic averaging method over $\mathcal G = SO(3)$. They apply an iteratively reweighted least squares (IRLS) procedure in the tangent space of $SO(3)$ to optimize different robust energy functions,  including the one that uses least absolute deviations. They claim that the use of the $\ell_{1/2}$ norm for deviations results in highest empirical accuracy. The empirical robustness of the two latter works is not theoretically guaranteed, even in simple settings. A recent deep learning method \cite{NeuroRA} solves a supervised version of rotation synchronization,
but it does not apply to the above unsupervised formulation of the problem.

\citet{truncatedLS} use least absolute deviations minimization for solving 1D translation synchronization, where $\mathcal G=\mathbb R$ with  addition. 
They propose a special version of IRLS and provide a deterministic exact recovery guarantee that depends on properties of the graph and its Laplacian. They do not explain their general result in an adversarial setting, but in a very special noisy setting.

Robustness results were established for least absolute deviations minimization in camera location estimation, which is somewhat similar to group synchronization  \cite{HandLV15,LUDrecovery}. These results assume special probabilistic setting, however, they have relatively weak assumptions on the corruption model.


 Several energy minimization solutions have been proposed to $SE(3)$ synchronization \cite{SE3_MCMC, SE3_SDP_jesus, SE3_Rosen, SE3_sync, SE3_RPCA}. This problem asks to jointly estimate camera rotations and locations from relative measurements of both. Neither of these solutions successfully address highly corrupted scenarios.

Other works on group synchronization, which do not minimize energy functions but aim to robustly recover corrupted solutions, 
screen corrupted edges using cycle consistency information. For a group $\mathcal{G}$ with group identity denoted by $e$, any $m \geq 3$, any cycle
$L=\{i_1i_2, i_2i_3\dots i_m i_1\}$ of length $m$ and any corresponding product of ground-truth group ratios along $L$, $g^*_L=g_{i_1i_2}^*g_{i_2i_3}^*\cdots g_{i_mi_1}^*$, the cycle-consistency constraint is 
$g^*_L= e$.
In practice, one is given the product of measurements, that is, $g_L=g_{i_1i_2}g_{i_2i_3}\cdots g_{i_mi_1}$, and in order to ``approximately satisfy the cycle-consistency constraint'' one tries to enforce $g_L$ to be sufficiently close to $e$.
\citet{Zach2010} uses the cycle-consistency constraint to detect corrupted relative rotations in $SO(3)$. It seeks to maximize a log likelihood function, which is based on the cycle-consistency constraint, using either belief propagation or convex relaxation. However, no theoretical guarantees are provided for the accuracy of outlier detection. Moreover, the log likelihood function implies very strong assumptions on the joint densities of the given relative rotations.
\citet{shen2016} classify the relative rotations as uncorrupted if they belong to any cycle that approximately satisfies the  cycle-consistency constraint. However, this work only exploits local information and cannot handle the adversarial corruption case, where corrupted cycles can be approximately consistent. 

An iterative reweighting strategy, IR-AAB \cite{AAB}, was proposed to detect and remove corrupted pairwise directions for the different problem of camera location estimation. It utilizes another notion of cycle-consistency to infer the corruption level of each edge. \citet{cemp} extend the latter idea, and interpret it as a message passing procedure, to solve group synchronization with any compact group. They refer to their new procedure as cycle-edge message passing (CEMP). While We follow ideas of \citet{cemp,AAB}, we directly solve for group elements, instead of estimating corruption levels,  using them to initial cleaning of edges and solving the cleaner problem with another method.

To the best of our knowledge,  the unified frameworks for group synchronization are \citet{ICMLphase,cemp,AMP_compact}. However, \citet{ICMLphase} and \citet{AMP_compact} assume special probabilistic models that do not address adversarial corruption. Furthermore, \citet{ICMLphase} only applies to Lie groups and the different setting of multi-frequencies.

\subsection{Contribution of This Work}
Current group synchronization solvers often do not perform well with highly corrupted and noisy group ratios. In order to address this issue, we propose a rotation synchronization solver that can address in practice high levels of noise and corruption. Our main ideas seem to generalize to group synchronization with any compact group, but more careful developments and testing are needed for other groups. We emphasize the following specific contributions of this work:
\begin{itemize}
\item
We propose the message passing least squares (MPLS) framework as an alternative paradigm to IRLS for group synchronization, and in particular, rotation synchronization. It uses the theoretically guaranteed CEMP algorithm 
for estimating the underlying corruption levels.
These estimates are then used for learning better weights for the weighted least squares problem.
\item 
We explain in Section \ref{sec:issue} why the common IRLS solver may not be accurate enough and in Section \ref{sec:mpls} why MPLS can overcome these obstacles. 
\item While MPLS can be formally applied to any compact group, we refine and test it for the group $\mathcal G = SO(3)$. We demonstrate state-of-the-art results for rotation synchronization with both synthetic data having nontrivial scenarios and real SfM data.


\end{itemize}

\section{Setting for Robust Group Synchronization}\label{sec:adversarial}
Some previous robustness theories for group synchronization typically assume a very special and often unrealistic corruption probabilistic model for very special groups \cite{deepti,wang2013exact}. In general, simplistic probabilistic models for corruption,
such as generating potentially corrupted group ratios according to the Haar measure on $\mathcal G$ \cite{wang2013exact},
may not generate some nontrivial scenarios that often occur in practice.  
For example, in the application of rotation synchronization that arise in SfM, the corrupted camera relative rotations can be self-consistent due to the ambiguous scene structures \citep{1dsfm14}. However, in common probabilistic models, such as the one in \citet{wang2013exact}, cycles with corrupted edges are self-consistent with probability zero. A more realistic model is the adversarial corruption model for the different problem of camera location \cite{LUDrecovery, HandLV15}. However, it also assumes very special probabilistic models for the graph and camera locations, which are not realistic. A more general model of adversarial corruption with noise is due to \citet{cemp} and we review it here.

We assume a graph $G([n],E)$ 
and a compact group $\mathcal G$
with a bi-invariant metric $d$, that is, for any $g_1$, $g_2$, $g_3\in \mathcal G$, 
 $d(g_1,g_2)=$ $d(g_3g_1,g_3g_2)=$ $d(g_1g_3,g_2g_3)$. 
 For $\mathcal G = SO(3)$, or any Lie group, $d$ is commonly chosen to be the geodesic distance.
 Since $\mathcal G$ is compact,  we can scale $d$ and assume 
 that $d(\cdot)\leq 1$.

We partition $E$ into $E_g$ and $E_b$, which represent sets of good (uncorrupted) and bad (corrupted) edges, respectively. 
We will need a topological assumption on $E_b$, or equivalently, $E_g$. A necessary assumption is that $G([n],E_g)$ is connected, though further restrictions on $E_b$ may be needed for 
establishing theoretical guarantees
\cite{cemp}.

In the noiseless case, the adversarial corruption model generates group ratios in the following way.
\begin{align}\label{eq:model}
g_{ij}=\begin{cases}
g^*_{ij}:=g_i^*g_j^{*-1}, & ij \in E_g;\\
\tilde g_{ij} \neq g^*_{ij}, & ij \in E_b.
\end{cases}
\end{align}
That is, for edges $ij\in E_b$, the corrupted group ratio $\tilde g_{ij}\neq g_{ij}^*$ can be arbitrarily chosen from $\mathcal G$.
The corruption is called adversarial since 
one can maliciously corrupt the group ratios for $ij\in E_b$ and also maliciously choose $E_b$ as long as the needed assumptions on $E_b$ are satisfied. One can even form cycle-consistent corrupted edges, so that they can be confused with the good edges.

In the noisy case, we assume a noise model for $d(g_{ij},g_{ij}^*)$, where $ij\in E_g$. In theory, one may need to restrict this model \cite{cemp}, but in practice we test highly noisy scenarios.

For $ij\in E$ we define the corruption level of $ij$ as \[s_{ij}^* = d(g_{ij},g_{ij}^*).\]
We use ideas of \citet{cemp} to estimate $\{s_{ij}^*\}_{ij \in E}$, but then we propose new ideas to estimate $\{g_{i}^*\}_{i \in [n]}$. While exact estimation of both quantities is equivalent in the noiseless case \cite{cemp}, this property is not valid when adding noise.

\section{Issues with the Common IRLS}\label{sec:issue}
We first review the least squares minimization, least absolute and unsquared deviations minimization and IRLS for group synchronization. We then explain why IRLS may not form a good solution for the group synchronization problem, and in particular for Lie algebraic groups, such as the rotation group. 

The least squares minimization 
can be formulated as follows:
\begin{align}
\label{eq:l2}
    \min_{\{g_i\}_{i=1}^n \subseteq \mathcal G}\sum_{ij\in E}d^2(g_{ij},g_ig_j^{-1}),
\end{align}
where one often relaxes this formulation. 
This formulation is generally sensitive to outliers and thus more robust energy functions are commonly used when considering corrupted group ratios. More specifically, one may choose a special function $\rho(x) \neq x^2$ and solve the following least unsquared deviation formulation
\begin{align}\label{eq:lp}
    \min_{\{g_i\}_{i=1}^n \subseteq \mathcal G}\sum_{ij\in E}\rho\left(d(g_{ij},g_ig_j^{-1})\right).
\end{align}
The special case of $\rho(x)=x$ \cite{HandLV15,HartleyAT11_rotation,ozyesil2015robust, wang2013exact} is referred to as least absolute deviations. Some other common choices are $\rho(x)=x^2/(x^2+\sigma^2)$ \cite{ChatterjeeG13_rotation} and $\rho(x)=\sqrt x$ \cite{L12}.

The least unsquared formulation is typically solved using IRLS, where  at iteration $t$ one solves the weighted least squares problem:
\begin{align} \label{eq:irls}
    \{g_{i,t}\}_{i\in [n]}&= \argmin_{\{g_i\}_{i=1}^n \subseteq \mathcal G}\sum_{ij\in E} w_{ij,t-1}d^2(g_{ij},g_ig_j^{-1}).
\end{align}
In the first iteration the weights can be initialized in a certain way, but in the next iterations the weights are updated using the residuals of this solution. Specifically, for $ij \in E$ and iteration $t$, the residual is $r_{ij,t}=d(g_{ij},g_{i,t}g_{j,t}^{-1})$ and the weight $w_{ij,t}$ is 
\begin{align}
w_{ij,t}&= 
F(r_{ij,t}),
\label{eq:weight_irls}
\end{align}
where the function $F$ depends on the choice of $\rho$. For $\rho(x)=x^p$, where $0<p<2$, $F(x)=\min\{x^{p-2},A\}$, where $1/A$ is a regularization parameter and here we fix  $A=10^8$.

The above IRLS procedure poses the following three issues. First, its convergence to the solution $\{g_i^*\}_{i\in [n]}$
is not guaranteed, especially under severe corruption. Indeed, IRLS succeeds when it accurately estimates the correct weights $w_{ij,t}$ for each edge. Ideally, when the solution $\{g_{i,t}\}_{i\in [n]}$ is close to the ground truth $\{g_i^*\}_{i\in [n]}$, the residual $r_{ij,t}$ must be close to the corruption level $s_{ij}^*$ so that weight  $w_{ij,t}$ must be close to $F(s_{ij}^*)$. 
However, if edge $ij \in E_b$ is severely corrupted (or edge $ij \in E_g$ has high noise) and either $g_i^*$ or $g_j^*$ is wrongly estimated, then the residual $r_{ij,t}$ might have a very small value. 
Thus the weight $w_{ij,t}$ in \eqref{eq:weight_irls} can be extremely large and may result in an inaccurate solution in the next iteration and possibly low-quality solution at the last iteration.

The second issue is that for common groups each iteration of \eqref{eq:irls} requires either SDP relaxation or tangent space approximation (for Lie groups).  However, if the weights of IRLS  are wrongly estimated in the beginning, then they may affect the tightness of the SDP relaxation and the validity of tangent space approximation.  Therefore, such procedures tend to make the IRLS scheme sensitive to corruption and initialization of weights and group elements.  

At last, when dealing with noisy 
data where most of $s_{ij}^*$, $ij \in E$, are significantly greater than $0$, the current reweighting strategy usually gives non-negligible positive weights to outliers. This can be concluded from the expression of $F$ (e.g., for $\ell_p$ minimization) and the fact that in a good scenario $r_{ij,t} \approx s_{ij}^*$ and  $s_{ij}^*$ can be away from 0. Therefore, outliers can be overweighed and this may lead to low-quality solutions. 
We remark that this issue is more noticeable in Lie groups, such as the rotation group, as all measurements are often noisy and corrupted; whereas in discrete groups some measurements may be rather accurate \cite{robust_multi_object2020}.

\section{Message Passing Least Squares (MPLS)}
\label{sec:mpls}
In view of the drawbacks of the common IRLS scheme, we 
propose the MPLS (Message Passing Least Squares), or Minneapolis, algorithm.
It carefully initializes and reevaluates the weights of a weighted least squares problem by our CEMP algorithm \cite{cemp} or a modified version of it. 
We first review the ideas of CEMP in Section \ref{sec:cemp}. We remark that its goal is to estimate the corruption levels $\{s_{ij}^*\}_{ij \in E}$ and not the group elements $\{g_{i}^*\}_{i \in [n]}$.  
Section \ref{sec:mpls2} formally describes MPLS for the general setting of group synchronization. Section \ref{sec:so3} carefully refines MPLS for rotation synchronization. Section \ref{sec:complexity} summarizes the complexity of the proposed algorithms.
\begin{figure*}[t]
    \centering
    \includegraphics[width=0.8\textwidth]{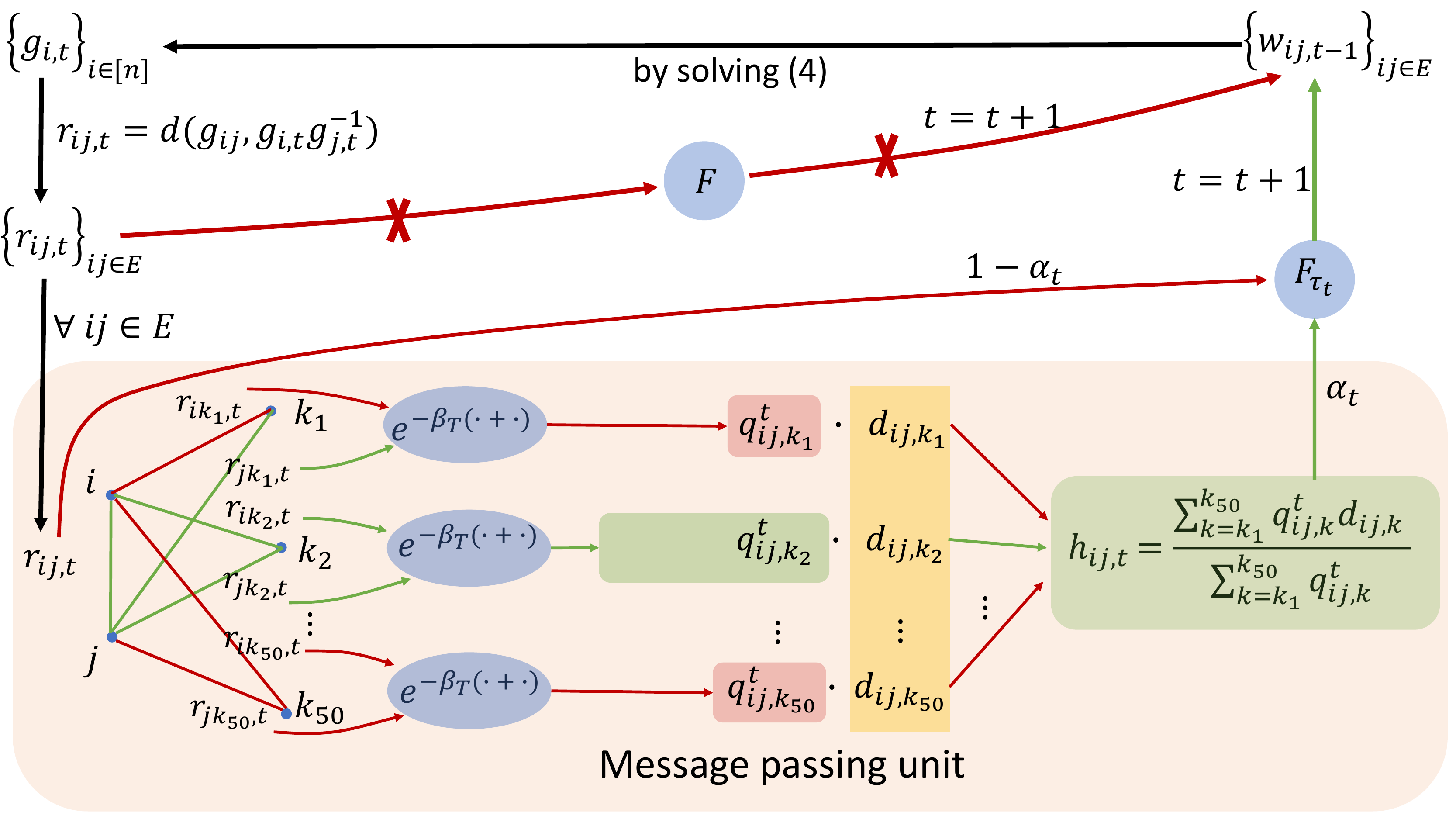}
    \caption{Demonstration of MPLS in comparison with IRLS. IRLS updates the graph weights directly from the residuals, which can be inaccurate. It is demonstrated in the upper loop of the figure, where the part different than MPLS is crossed out. In contrast, MPLS updates the graph weights by applying a CEMP-like procedure to the residuals, demonstrated in the ``message-passing unit''. Good edges, such as  $jk_1$, are marked with green, and bad edges are marked with red. For $ij\in E$ and $k\in \{k_1,k_2,\dots k_{50}\}$, $q_{ij,k}^t$ is updated using the two residuals $r_{ik,t}$ and $r_{jk,t}$ according to the indicated operation. The length of a bar around the computed value of each $q_{ij,k}^t$ is proportional to magnitude and the green or red colors designate good or bad corresponding cycles, respectively. 
    The weighted sum $h_{ij,t}$ aims to approximate $s_{ij}^*$ and this approximation is good when the green $q_{ij,k}^t$ bars  are much longer than the red bars. The weight $w_{ij,t}$ is formed as a convex combination of $r_{ij,t}$ and $h_{ij,t}$. The rest of the procedure is similar to IRLS. }\label{fig:demo}
\end{figure*}

\subsection{Cycle-Edge Message Passing (CEMP)}
\label{sec:cemp}
The CEMP procedure aims to estimate the corruption levels
$\{s_{ij}^*\}_{ij \in E}$
from the cycle inconsistencies, which we define next. 
For simplicity and for ease of computation, we work here with 3-cycles, that is, triangles in the graph. 
For each edge $ij\in E$,  we independently sample with replacement 50 
nodes that form 3-cycles with $i$ and $j$. That is, if $k$ is such a node then $ik$ and $jk$ are in $E$.
We denote this set of nodes by $C_{ij}$. We remark that the original version of CEMP in \citet{cemp} uses all 3-cycles and can be slower.
We define the cycle inconsistency of the 3-cycle $ijk$ associated with edge $ij$ and $k \in C_{ij}$ as follows
\begin{equation}
\label{eq:def_dL}
d_{ij,k} := d(g_{ij}g_{jk}g_{ki}, e), \quad k\in C_{ij}, ij\in E.
\end{equation}

The idea of CEMP is to iteratively estimate each corruption level  $s_{ij}^*$ for $ij \in E$ from a weighted average of the cycle inconsistencies $\{d_{ij,k}\}_{k \in C_{ij}}$. 
To motivate this idea, we assume the noiseless adversarial corruption model
and formulate the following proposition whose easy proof appears in \citet{cemp}.
\begin{proposition}\label{prop:good cycle}
If $s_{ik}^*=s_{jk}^*=0$, that is, $ik$, $jk \in E_g$, then $$s_{ij}^*=d_{ij,k}.$$
\end{proposition}
If the condition of the above proposition holds, we call $ijk$ a good cycle with respect to $ij$, 
otherwise we call it a bad cycle.

CEMP also aims to estimate the conditional probability $p_{ij,k}^t$ that the cycle $ijk$ is good, so $s_{ij}^*=d_{ij,k}.$ The conditioning is on the estimates of corruption levels computed in the previous iteration. CEMP uses this probability as a weight for the cycle inconsistency $d_{ij,k}$. The whole weighted sum
thus aims to estimate the conditional expectation of $s_{ij}^*$ at iteration $t$. This estimate is denoted by $s_{ij,t}$.
The iteration of CEMP thus contains two main steps:
1) Computation of the weight $p_{ij,k}^t$ of $d_{ij,k}$; 2) Computation of $s_{ij,t}$ as a weighted average of the cycle inconsistencies $\{d_{ij,k}\}_{k \in C_{ij}}$. In the former stage messages are passed from edges to cycles and in the latter stage messages are passed from cycles to edges. The simple procedure is summarized in Algorithm~\ref{alg:cemp}. We formulate it in generality for a compact group $\mathcal G$ with 
a bi-invariant metric $d$ 
and a graph $G([n],E)$, for which the cycle inconsistencies, $\{d_{ij,k}\}_{ij\in E, k\in C_{ij}}$, were computed in advance.
Our default parameters are later specified in Section \ref{sec:implementation}. For generality, we write $|C_{ij}|$ instead of 50. \newpage
\begin{algorithm}[h]
\caption{CEMP \cite{cemp}}
\label{alg:cemp}
\begin{algorithmic}
\REQUIRE $\{d_{ij,k}\}_{ij\in E, k\in C_{ij}}$, time step $T$, increasing  $\{\beta_t\}_{t=0}^T$\\
\STATE \textbf{Steps:}
\STATE $s_{ij,0} = \frac{1}{|C_{ij}|}\sum_{k\in C_{ij}} d_{ij,k} $ \hspace*{\fill} $ij\in E$
\FOR {$t=0:T$}
\STATE \emph{Message passing from edges to cycles:}
\STATE $p_{ij,k}^{t} =  \exp\left(-\beta_t(s_{ik,t}+s_{jk,t})\right)$ \hspace*{\fill} $k\in C_{ij},\, ij\in E$
\STATE \emph{Message passing from cycles to edges:}
\STATE $s_{ij,t+1}
    =\frac{1}{Z_{ij,t}}\sum\limits_{k\in C_{ij}}p_{ij,k}^t d_  {ij,k}$,   
\hspace*{\fill} $ij\in E$
\STATE
where $Z_{ij,t}=\sum\limits_{k\in C_{ij}}p_{ij,k}^t$ is a normalization factor
\ENDFOR
\ENSURE $\{s_{ij,T}\}_{ij\in E}$
\end{algorithmic}
\end{algorithm}

Algorithm \ref{alg:cemp} can be explained in two different ways \cite{cemp}. First of all, it can be theoretically guaranteed to be robust to adversarial corruption and stable to low level of noise (see Theorem 5.4 in \citet{cemp}). Second of all, our above heuristic explanation can be made more rigorous using some statistical assumptions. Such assumptions are common in other message passing algorithms in statistical mechanics formulations and we thus find it important to motivate them here. We remark that our statistical assumptions are weaker than those of previous works on message passing \cite{AMP_Donoho,BP}, and in particular, those for group synchronization \cite{AMP_compact,Zach2010}. We also remark that they are not needed for establishing the above mentioned theory.

Our first assumption is that $ijk$ is a good cycle if and only if $s_{ij}^*=d_{ij,k}$. Proposition \ref{prop:good cycle} implies the only if part, but the other part is generally not true. However, under special random corruption models (e.g., models in \citet{wang2013exact,ozyesil2015robust}), the assumed equivalence holds with probability $1$. 
We further assume that $\{s_{ij}^*\}_{ij \in E}$ and $\{s_{ij,t}\}_{ij \in E}$ are both i.i.d.~random variables and that for any $ij \in E$, $s_{ij}^{*}$ is independent of $s_{kl,t}$ for $kl \neq ij \in E$. We further assume that for any $ij\in E$
\begin{equation}
\label{eq:pr_f}
\Pr(s_{ij}^*=0|s_{ij,t}=x) = \exp(-\beta_t x).
\end{equation}
We also assume the existence of good cycle $ijk$ for any $ij\in E$. 

In view of these assumptions, in particular the i.i.d.~sampling and \eqref{eq:pr_f}, we obtain that the expression for $p_{ij,k}^t$ used in Algorithm \ref{alg:cemp} coincides with the conditional probability that $ijk$ is a good cycle, that is, the conditional probability that $s_{ik}^*=s_{jk}^*=0$:
\begin{align}\label{eq:pijk}
    p_{ij,k}^t &= \Pr(s_{ik}^*=s_{jk}^*=0|\{s_{ab,t}\}_{ab\in E})\nonumber\\
    &= \Pr(s_{ik}^*=0|s_{ik,t})\Pr(s_{jk}^*=0|s_{jk,t})\\ \nonumber
    &=\exp(-\beta_t (s_{ik,t} +s_{jk,t})).
\end{align}

Using the definition of conditional expectation, 
 the equivalence assumption, the above i.i.d.~sampling assumptions and \eqref{eq:pr_f},
 we show that the expression for $s_{ij,t}$ used in Algorithm \ref{alg:cemp} coincides with the conditional  expectation of $s_{ij}^*$:
 \begin{align}\label{eq:sijt}
     &\mathbb E\left(s_{ij}^*|\{s_{ab,t}\}_{ab\in E}\right)\nonumber\\
     &=\frac{1}{Z_{ij,t}}\sum_{k\in C_{ij}}\Pr\left(s_{ij}^*=d_{ij,k}|\{s_{ab,t}\}_{ab\in E}\right)d_{ij,k}\nonumber
      \end{align}
     \begin{align}
     &=\frac{1}{Z_{ij,t}}\sum_{k\in C_{ij}}\Pr\left(s_{ik}^*=s_{jk}^*=0|\{s_{ab,t}\}_{ab\in E}\right)d_{ij,k} \nonumber\\
     &=\frac{1}{Z_{ij,t}}\sum_{k\in C_{ij}}\exp\left(-\beta_t(s_{ik,t}+s_{jk,t})\right)d_{ij,k}
     \\     &= 
     \frac{1}{Z_{ij,t}}\sum_{k\in C_{ij}} p_{ij,k}^t d_{ij,k}.\nonumber
 \end{align}
Note that our earlier motivation of  Algorithm
 \ref{alg:cemp} assumed both
 \eqref{eq:pijk} and \eqref{eq:sijt}.
Demonstration of a procedure similar to CEMP, but with different notation, appears in the lower part of Figure \ref{fig:demo}.


\subsection{General Formulation of MPLS}
\label{sec:mpls2}
MPLS uses the basic ideas of CEMP in order to robustly estimate the residuals $\{r_{ij,t}\}_{ij \in E}$ and weights $\{w_{ij,t}\}_{ij \in E}$ of the IRLS scheme as well as to carefully initialize this scheme. It also incorporates a novel truncation idea. We explain in this and the next section how these new ideas address the drawbacks of the common IRLS procedure, which was reviewed in Section \ref{sec:issue}.
We sketch MPLS in Algorithm \ref{alg:MPLS}, demonstrate it in Figure \ref{fig:demo} and explain it below. For generality, we assume in this section a compact group $\mathcal G$, 
a bi-invariant metric $d$ on $\mathcal G$ and a graph $G([n],E)$ with given relative measurements and cycle inconsistencies computed in advance. 
Our default parameters are later specified in Section \ref{sec:implementation}.

Instead of using the traditional IRLS reweighting function $F(x)$ explained in Section \ref{sec:issue}, we use its truncated version $F_{\tau}(x)=F(x)\mathbf{1}_{\{x\leq \tau\}}+10^{-8}\mathbf{1}_{\{x> \tau\}}$ with a parameter $\tau>0$. We decrease $\tau$ as the  iteration number increases in order to  avoid overweighing outliers. By doing this we aim to address the third drawback of IRLS mentioned in Section \ref{sec:issue}. We remark that the truncated function is $F(x)\mathbf{1}_{\{x\leq \tau\}}$ and the additional term $10^{-8}\mathbf{1}_{\{x> \tau\}}$ is needed to ensure that the graph with weights resulting from $F_{\tau}$ is connected.



\begin{algorithm}[h]
\caption{Message Passing Least Squares (MPLS)}\label{alg:MPLS}
\begin{algorithmic}
\REQUIRE $\{g_{ij}\}_{ij\in E}$, $\{d_{ij,k}\}_{k\in C_{ij}}$, nonincreasing $\{\tau_t\}_{t\geq 0}$, increasing  $\{\beta_t\}_{t=0}^T$, decreasing $\{\alpha_t\}_{t\geq 1}$
\STATE \textbf{Steps:}
\STATE Compute $\{s_{ij,T}\}_{ij\in E}$ by CEMP
\STATE $w_{ij,0}=F_{\tau_0}(s_{ij,T})$
\hspace*{\fill} $ij\in E$
\STATE $t=0$
\WHILE {not convergent}
\STATE $t=t+1$
\STATE $\{g_{i,t}\}_{i\in [n]}=\argmin\limits_{g_i\in\mathcal G}\sum\limits_{ij\in E} w_{ij,t-1}d^2(g_{ij},g_ig_j^{-1})$

\STATE $r_{ij,t}=d(g_{ij}, g_{i,t}g_{j,t}^{-1})$
\hspace*{\fill} $ij\in E$

\STATE $q_{ij,k}^{t} = \exp(-\beta_T(r_{ik,t}+r_{jk,t}))$
\hspace*{\fill} $k\in C_{ij},\, ij\in E$

\STATE $h_{ij,t}
    =\frac{\sum_{k\in C_{ij}}q_{ij,k}^t d_  {ij,k}}{\sum_{k\in C_{ij}}q_{ij,k}^t}$
\hspace*{\fill} $ij\in E$
\STATE $w_{ij,t}=F_{\tau_t}(\alpha_t h_{ij,t}+(1-\alpha_t)r_{ij,t})$
\hspace*{\fill} $ij\in E$
\ENDWHILE
\ENSURE $\left\{g_{i,t}\right\}_{i\in [n]}$
\end{algorithmic}
\end{algorithm}
The initial step of the algorithm estimates the corruption levels $\{s_{ij}^*\}_{ij \in E}$ by CEMP. The initial weights for the IRLS procedure 
follow \eqref{eq:weight_irls} with additional truncation. 
At each iteration, the group ratios $\{g_{i,t}\}_{i\in [n]}$ are estimated from the weighted least squares procedure in \eqref{eq:irls}. However, the weights $w_{ij,t}$ are updated in a very different way. First of all, for each $ij \in E$ the corruption level $s_{ij}^*$ is re-estimated in two different ways and a convex combination of the two estimates is taken. The first estimate is a residual $r_{ij,t}$ computed with the newly updated estimates $\{g_{i,t}\}_{i\in [n]}$. This is the error of approximating the given measurement $g_{ij}$ by the newly estimated group ratio. The other estimate practically applies CEMP to re-estimate the corruption levels. For edge $ij \in E$, the latter estimate of $s_{ij}^*$ is denoted by $h_{ij,t}$. 
For interpretation, we can replace \eqref{eq:pr_f} with 
$\Pr(s_{ij}^*|r_{ij,t})=\exp(-\beta_T x)$ and use it to derive analogs of \eqref{eq:pijk} 
and 
\eqref{eq:sijt}.
Unlike CEMP, we use the single parameter, $\beta_T$, as we assume that CEMP provides a sufficiently good initialization.
At last, a similar weight as in \eqref{eq:weight_irls}, but truncated, is applied to the combined estimate $\alpha_t h_{ij,t}+(1-\alpha_t)r_{ij,t}$.

We remark that utilizing the estimate $h_{ij,t}$ for the corruption level addresses the first drawback of IRLS discussed in Section \ref{sec:issue}. Indeed, assume the case where $ij\in E_b$ and  $r_{ij,t}$ is close to 0. Here, $w_{ij,t}$ computed by IRLS is relatively large; however, since $ij\in E_b$, $w_{ij,t}$ needs to be small. Unlike $r_{ij,t}$ in IRLS, we expect that $h_{ij,t}$ in MPLS should not be too small as long as for some $k\in C_{ij}$, $d_{ij,k}$ are sufficiently large. This happens as long as there exists some $k\in C_{ij}$ for which the cycle $ijk$ is good. Indeed, in this case $s_{ij}^*$ is sufficiently large and for good cycles $d_{ij,k}=s_{ij}^*$.

We further remark that $h_{ij,t}$ is a good approximation of $s_{ij}^*$ under certain conditions. For example, if for all $k\in C_{ij}$, $r_{ik,t}\approx s_{ik}^*$ and $r_{jk,t}\approx s_{jk}^*$, then plugging in the definition of $p_{ij,k}^t$ to the expression of $h_{ij,t}$, using the fact that $\beta_T$ is sufficiently large and at last applying Proposition \ref{prop:good cycle}, we obtain that
\begin{align}
    h_{ij,t}
    &= \sum_{k\in C_{ij}}\frac{\exp(-\beta_T(r_{ik,t}+r_{jk,t}))}{\sum_{k\in C_{ij}}\exp(-\beta_T(r_{ik,t}+r_{jk,t}))}d_  {ij,k} \nonumber\\
    &\approx \sum_{k\in C_{ij}}\frac{\exp(-\beta_T(s_{ik}^*+s_{jk}^*)) }{\sum_{k\in C_{ij}}\exp(-\beta_T(s_{ik}^*+s_{jk}^*))} d_  {ij,k} \\
    &\approx \sum_{k\in C_{ij}}\frac{\mathbf{1}_{\{ijk \text{ is a good cycle}\}}}{\sum_{k\in C_{ij}}\mathbf{1}_{\{ijk \text{ is a good cycle}\}}} d_  {ij,k}=s_{ij}^*.\nonumber
\end{align}
This intuitive argument for a restricted case conveys the idea that ``local good information'' can be used to estimate $s_{ij}^*$. 
The theory of CEMP \cite{cemp} shows that under weaker conditions such information can propagate through the whole graph within a few iterations, but we cannot extend it to MPLS. 

If the graph $G([n], E)$ is dense with sufficiently many good cycles, then we expect that this good information can propagate in few iterations and that $h_{ij,t}$ will have a significant advantage over $r_{ij,t}$. However, in real scenarios of rotation synchronization in SfM, one may encounter sparse graphs, which may not have enough cycles and, in particular, not enough good cycles. In this case, utilizing $h_{ij,t}$ is mainly useful in the early iterations of the algorithm. On the other hand, when $\{g_{i,t}\}_{i \in [n]}$ are close to $\{g_i^*\}_{i \in [n]}$,  $\{r_{ij,t}\}_{i \in [n]}$ will be sufficiently close to $\{s_{ij}^*\}_{i \in [n]}$. Aiming to address rotation synchronization, we decrease $\alpha_t$, the weight of $h_{ij,t}$,  with $t$. In other applications, different choices of $\alpha_t$ can be used \cite{robust_multi_object2020}.

The second drawback of IRLS, discussed in Section \ref{sec:issue}, is the possible difficulty of implementing the weighted least squares step of \eqref{eq:irls}. This issue is application-dependent, and since in this work we focus on rotation  synchronization (equivalently, $SO(3)$ synchronization), we show in the next subsection how MPLS can deal with the above issue in this specific problem. 
Nevertheless, we claim that our framework can also be applied to other compact group synchronization problems and we demonstrate this claim in a follow up work \cite{robust_multi_object2020}.

\subsection{MPLS for $SO(3)$ synchronization}  
\label{sec:so3}
Rotation synchronization, or $SO(3)$ synchronization, aims to solve 3D rotations $\{\bR_i^*\}_{i\in [n]} \in SO(3)$ from measurements $\{\bR_{ij}\}_{ij\in E} \in SO(3)$ of the 3D relative rotations  $\{\bR_i^*\bR_j^{*-1}\}_{ij\in E} \in SO(3)$. 
Throughout the rest of the paper, we use the following normalized geodesic distance  for  $\bR_1,\bR_2 \in SO(3)$:
\begin{equation}
\label{eq:geo_distance}
 d(\bR_1,\bR_2)=\|\log(\bR_1\bR_2^{-1})\|_F/(\sqrt2\pi),
\end{equation}
where $\log$ is the matrix logarithm and the normalization factor ensures that the diameter of $SO(3)$ is $1$. 
We provide some relevant preliminaries of the Riemannian geometry of $SO(3)$ in Section \ref{sec:prelim_riem} and then describe the implementation of MPLS for $SO(3)$, which we refer to as MPLS-$SO(3)$, in Section \ref{sec:mpls_so3_details}.

\subsubsection{Preliminaries: $SO(3)$ and  $\mathfrak{so}(3)$} \label{sec:prelim_riem}
We note that $SO(3)$ is a Lie group, and its corresponding Lie algebra, $\mathfrak{so}(3)$, is the space of all skew symmetric matrices, which is isomorphic to $\mathbb R^3$. For each $\bR\in SO(3)$, its corresponding element in $\mathfrak{so}(3)$ is $\boldsymbol\Omega=\log(\bR)$, where $\log$ denotes matrix logarithm. Each $\boldsymbol\Omega\in \mathfrak{so}(3)$ can be  represented as  $[\boldsymbol{\omega}]_{\times}$ for some $\boldsymbol\omega =(\omega_1, \omega_2, \omega_3)^T \in \mathbb R^3$ in the following way:
\begin{align*}
    [\boldsymbol{\omega}]_{\times} :=
    \left(\begin{array}{ccc}
        0 & - \omega_3 & \omega_2 \\
        \omega_3 & 0&  -\omega_1  \\
        -\omega_2 & \omega_1 & 0
    \end{array}\right).
\end{align*}
In other words, we can map any $\boldsymbol\omega\in \mathbb R^3$ to 
$\boldsymbol\Omega=[\boldsymbol{\omega}]_{\times} \in \mathfrak{so}(3)$ and $\bR=\exp([\boldsymbol{\omega}]_{\times}) \in SO(3)$, where $\exp$ denotes the matrix exponential function. We remark that geometrically $\boldsymbol\omega$ is the tangent vector at $\boldsymbol I$ of the geodesic path from $\boldsymbol I$ to $\bR$.

\subsubsection{Details of MPLS-$SO(3)$}
\label{sec:mpls_so3_details}
We note that in order to adapt MPLS to the group $SO(3)$, we only need a specific algorithm to solve the following  formulation of the weighted least squares problem at iteration $t$
\begin{align}
   \nonumber &\min\limits_{\bR_{i,t}\in \mathcal SO(3)}\sum\limits_{ij\in E} w_{ij,t}d^2(\bR_{ij},\bR_{i,t}\bR_{j,t}^{-1})\\
    =& \min\limits_{\bR_{i,t}\in \mathcal SO(3)}\sum\limits_{ij\in E} w_{ij,t}d^2(\boldsymbol{I}, \bR_{i,t}^{-1}\bR_{ij}\bR_{j,t}),\label{eq:wlsSO3}
\end{align}
where the last equality follows from the bi-invariance of $d$.
The constraints on orthogonality and determinant of $\bR_i$ are non-convex. If one relaxes those constraints, with an appropriate choice of the metric $d$, then the solution of the least squares problem in the relaxed Euclidean space often lies away from the embedding of $SO(3)$ into that space. 
For this reason, we follow the common choice of $d$ according to \eqref{eq:geo_distance} and implement the Lie-algebraic Averaging (LAA) procedure \cite{Govindu04_Lie, ChatterjeeG13_rotation, L12,consensusSO3}. 
We review LAA, explain why it may be problematic and why our overall implementation may overcome its problems. 
LAA aims to move from  $\bR_{i,t}$ to $\bR_{i,t+1}$ along the manifold using the right group action $\bR_{i,t}=\bR_{i,t-1}\Delta \bR_{i,t}$, where 
$\Delta \bR_{i,t}\in SO(3)$. 
For this purpose, it defines $\Delta\bR_{ij,t}=\bR_{i,t-1}^{-1}\bR_{ij}\bR_{j,t-1}$
so that 
\begin{multline*}
(\Delta \bR_{i,t})^{-1}\Delta \bR_{ij,t} \Delta \bR_{j,t} = \\(\Delta \bR_{i,t})^{-1}\bR_{i,t-1}^{-1}\bR_{ij}\bR_{j,t-1} \Delta \bR_{j,t}=\bR_{i,t}^{-1}\bR_{ij}\bR_{j,t}    
\end{multline*}
and \eqref{eq:wlsSO3} 
can be transformed to the still hard to solve equation 
\begin{equation}\label{eq:deltaR}
    \min\limits_{\Delta \bR_{i,t}\in\mathcal SO(3)}\sum\limits_{ij\in E} w_{ij,t}d^2(\boldsymbol I,(\Delta \bR_{i,t})^{-1}\Delta \bR_{ij,t} \Delta \bR_{j,t}).
\end{equation}
LAA then maps $\{\Delta \bR_{i,t}\}_{i\in [n]}$ and $\{\Delta \bR_{ij,t}\}_{ij\in E}$ to the tangent space of $\boldsymbol{I}$ by $\Delta \boldsymbol\Omega_{i,t}=\log \Delta \bR_{i,t}$
and $\Delta \boldsymbol\Omega_{ij,t}=\log \Delta \bR_{ij,t}$.
Applying \eqref{eq:geo_distance} and the fact that the Riemannian logarithmic map, which is represented by $\log$, preserves the geodesic  distance and using a ``naive approximation'':
$d(\boldsymbol I,(\Delta \bR_{i,t})^{-1}\Delta \bR_{ij,t} \Delta \bR_{j,t}) $. Therefore, 
LAA uses the following approximation  
\begin{multline}
\label{eq:app_laa}
 d(\boldsymbol{I},(\Delta \bR_{i,t})^{-1}\Delta \bR_{ij,t}\Delta \bR_{j,t}) 
 = \\
 \|\log((\Delta \bR_{i,t})^{-1}\Delta \bR_{ij,t} \Delta \bR_{j,t})\|_F/(\sqrt2\pi)
 \approx \\
\|-\log(\Delta \bR_{i,t})+\log(
\Delta \bR_{ij,t})+\log( \Delta \bR_{j,t}))\|_F/(\sqrt2\pi) = \\
\|\Delta \boldsymbol\Omega_{i,t}-\Delta \boldsymbol\Omega_{j,t}-\Delta \boldsymbol\Omega_{ij,t}\|_F/(\sqrt2\pi).
\end{multline}
Consequently, LAA transforms \eqref{eq:deltaR} as follows:
\begin{equation}\label{eq:delta_omega}
    \min\limits_{\Delta \boldsymbol\Omega_{i,t}\in\mathfrak{so}(3)}\sum\limits_{ij\in E} w_{ij,t}\|\Delta \boldsymbol\Omega_{i,t}-\Delta \boldsymbol\Omega_{j,t}-\Delta \boldsymbol\Omega_{ij,t}\|^2_F.
\end{equation}
However, the approximation in \eqref{eq:app_laa}
is only valid when $\Delta \bR_{ij,t}$, $\Delta \bR_{i,t}$, $\Delta \bR_{j,t}$  $\approx \boldsymbol{I}$, 
which is unrealistic. 

One can check that the following conditions: $\bR_{ij}\approx \bR_i^*\bR_j^{*-1}$ ($s_{ij}^*\approx 0$), $\bR_{i,t} \approx \bR_i^*$ and $\bR_{j,t} \approx \bR_j^*$ for $t \geq 0$ imply that $\Delta \bR_{ij,t}$, $\Delta \bR_{i,t}$, $\Delta \bR_{j,t}$  $\approx \boldsymbol{I}$ and thus  
imply \eqref{eq:app_laa}. Therefore, to make LAA work we need to give large weights to edges $ij$ with small $s_{ij}^*$ and provide a good initialization $\{\bR_{i,0}\}_{i\in [n]}$ that is reasonably close to $\{\bR_{i}^*\}_{i\in [n]}$ and so that $\{\bR_{i,t}\}_{i\in [n]}$ for all $t\geq 1$ are still close to the ground truth. 
Our heuristic argument is that good approximation by CEMP, followed by MPLS, addresses these requirements. Indeed, to address the first requirement, we note that good initialization by CEMP can result in $s_{ij,T} \approx s_{ij}^*$ and by the nature of $F$,  $w_{ij,0}$ is large when $s_{ij,T}$ is small. 
As for the second requirement, 
we assign the weights $s_{ij,T}$, obtained by CEMP, to each $ij\in E$ and find the minimum spanning tree (MST) for the weighted graph by Prim's algorithm. We initialize the rotations by fixing $\bR_{1,0}=\boldsymbol{I}$, multiplying relative rotations along the computed MST and consequently obtaining $\bR_{i,0}$ for any node $i$. We summarize our MPLS version of rotation averaging in Algorithm~\ref{alg:SO3}. 
\begin{algorithm}[h]
\caption{MPLS-$SO(3)$}\label{alg:SO3}
\begin{algorithmic}
\REQUIRE $\{\bR_{ij}\}_{ij\in E}$, $\{d_{ij,k}\}_{k\in C_{ij}}$, $\{\tau_t\}_{t\geq 0}$, $\{\beta_t\}_{t=0}^T$, $\{\alpha_t\}_{t\geq 1}$
\STATE \textbf{Steps:}

\STATE  Compute $\{s_{ij,T}\}_{ij\in E}$ by CEMP
\STATE Form an $n \times n$ weight matrix $\bW$, where $W_{ij}=W_{ji}= s_{ij,T}$ for $ij\in E$, and $W_{ij}=W_{ji}=0$ otherwise
\STATE $G([n],E_{ST})=$ minimum spanning tree of $G([n],W)$
\STATE $\bR_{1,0}=\boldsymbol{I}$
\STATE find $\{\bR_{i,0}\}_{i>1}$ by $\bR_i=\bR_{ij}\bR_j$ for $ij\in E_{ST}$ 
\STATE $t=0$
\STATE $w_{ij,0}=F_{\tau_0}(s_{ij,T})$ 
\WHILE {not convergent}
\STATE $t=t+1$
\STATE $\Delta \boldsymbol\Omega_{ij,t}=\log(\bR_{i,t-1}^{-1}\bR_{ij}\bR_{j,t-1})$ \hspace*{\fill} $ij\in E$

\STATE $\{\Delta \boldsymbol\Omega_{i,t}\}_{i\in [n]}=$
\STATE \quad \quad $\argmin\limits_{\Delta \boldsymbol\Omega_{i,t}\in\mathfrak{so}(3)}\sum\limits_{ij\in E} w_{ij,t}\|\Delta \boldsymbol\Omega_{i,t}-\Delta \boldsymbol\Omega_{j,t}-\Delta \boldsymbol\Omega_{ij,t}\|^2_F$
\STATE $\bR_{i,t}=\bR_{i,t-1}\exp(\Delta \boldsymbol\Omega_{i,t})$ \hspace*{\fill} $i\in [n]$
\STATE $r_{ij,t}=\|\Delta \boldsymbol\Omega_{i,t}-\Delta \boldsymbol\Omega_{j,t}-\Delta \boldsymbol\Omega_{ij,t}\|_F/(\sqrt2\pi)$ \hspace*{\fill} $ij\in E$

\STATE $q_{ij,k}^{t} =\exp(-\beta_T(r_{ik,t}+r_{jk,t}))$
\hspace*{\fill} $k\in C_{ij},\, ij\in E$
\STATE $h_{ij,t}
    =\frac{\sum_{k\in C_{ij}}q_{ij,k}^t d_  {ij,k}}{\sum_{k\in C_{ij}}q_{ij,k}^t}$ \hspace*{\fill} $ij\in E$
\STATE $w_{ij,t}=F_{\tau_t}(\alpha_t h_{ij,t}+(1-\alpha_t)r_{ij,t})$
\hspace*{\fill} $ij\in E$
\ENDWHILE
\ENSURE $\left\{\bR_{i,t}\right\}_{i\in [n]}$
\end{algorithmic}
\end{algorithm}

\subsection{Computational Complexity}
\label{sec:complexity}
CEMP requires the computation 
of $d_{ij,k}$ for $ij \in E$ and $k\in C_{ij}$. Its computational complexity per iteration is thus of order $O(|E|)$ as we use $|C_{ij}|=50$ for all $ij \in E$. Since we advocate few iterations ($T=5$) of CEMP, or due to its fast convergence under special settings \cite{cemp}, we can assume that its total complexity is $O(|E|)$.
The computational complexity of MPLS depends on the complexity of solving the weighted least squares problem, which depends on the group. For MPLS-$SO(3)$, the most expensive part is solving the weighted least squares problem in the tangent space, whose complexity is at most $O(n^3)$. This is thus also the complexity of MPLS-$SO(3)$ per iteration. Unlike CEMP, we have no convergence guarantees yet for MPLS.

\section{Numerical Experiments}
\label{sec:numerics}
We test the proposed MPLS algorithm on rotation synchronization, while comparing with state-of-the-art methods. We also try simpler ideas than MPLS that are based on the basic strategy of CEMP.
All computational tasks were implemented on a machine with 2.5GHz Intel i5 quad core processors and 8GB memory. 

\subsection{Implementation}
\label{sec:implementation}
We use the following default parameters for Algorithm \ref{alg:cemp}: 
$|C_{ij}|=50$ for $ij\in E$; $T=5$; $\beta_t=2^{t}$ and $t=0, \ldots, 5$. If an edge is not contained in any 3-cycle, we set its corruption level as 1. 
For MPLS-$SO(3)$, which we refer to in this section as MPLS, we use the above parameters of Algorithm \ref{alg:cemp} and the following ones for $t\geq 1$: 
 $$\alpha_t=1/(t+1) \ \text{ and } \ \tau_{t}=\inf_x\left\{\hat P_t(x)> \max\{1-0.05t\,,0.8\}\right\}.$$
Here, $\hat P_t$ denotes the empirical distribution of 
$\{\alpha_t h_{ij,t}+$ $(1-\alpha_t)r_{ij,t}\}_{ij\in E}$.
That is, for $t=0$, 1, 2, 3, we ignore $0\%$, $5\%$, $10\%$, $15\%$ of edges that have highest $\alpha_t h_{ij,t}+$ $(1-\alpha_t)r_{ij,t}$, and for $t \geq 4$ we ignore $20\%$ of such edges. 
$F(x)$ for MPLS is chosen as $x^{-3/2}$ and it corresponds to $\rho(x)=\sqrt x$. For simplicity and consistency, we use these choices of parameters for all of our experiments. We remark that our choice of $\beta_t$ in Algorithm \ref{alg:cemp} is supported by the theory of \citet{cemp}. 
We found that MPLS is not so sensitive to its parameters. One can choose other values of $\{\beta_t\}_{t\geq 0}$, for example any geometric sequence with ratio 2 or less, and stop after several iterations. Similarly, one may replace 0.8 and 0.05 
in the definition of $\tau_t$ with $0.7-0.9$ and $0.01-0.1$, respectively, and perform similarly on average.

We test two previous state-of-the-art IRLS methods: IRLS-GM \cite{ChatterjeeG13_rotation} with $\rho(x) = x^2/(x^2+25)$,  $F(x)=25/(x^2+25)^2$
and IRLS-$\ell_{1/2}$ \cite{L12} with $\rho(x)=\sqrt x$,  $F(x)=x^{-3/2}$.
We use their implementation by \citet{L12}. 

We have also separately implemented the part of initializing the rotations 
of MPLS in Algorithm~\ref{alg:SO3} and refer to it by CEMP+MST. Recall that it solves rotations by direct propagation along the minimum weighted spanning tree of the graph with weights obtained by 
Algorithm \ref{alg:cemp} (CEMP).
We also test the application of this initialization to  
the main algorithms in  \citet{ChatterjeeG13_rotation} and \citet{L12} and refer to the resulting methods by CEMP+IRLS-GM
and CEMP+IRLS-$\ell_{1/2}$, respectively.  We remark that the original algorithms initialize by a careful least absolute deviations minimization.
We use the convergence criterion
$\sum_{i\in[n]}\|\Delta \boldsymbol\Omega_{i,t}\|_F/(\sqrt2n)< 0.001$
of \citet{L12} for all the above algorithms.

Because the solution is determined up to a right group action, we align our estimated rotations $\{\hat\bR_i\}$ with the ground truth ones $\{ \bR_i^*\}$. That is, we find a rotation matrix  $\bR_\text{align}$ so that $\sum_{i\in [n]}\|\hat\bR_i\bR_\text{align}-\bR_i^*\|_F^2$ is minimized. For synthetic data, we report the following mean estimation error in degrees: $180\cdot\sum_{i\in [n]} d(\hat \bR_i\bR_\text{align}\,, \bR_i^*)/n$. For real data, we also report the median of $\{180\cdot d(\hat \bR_i\bR_\text{align}\,, \bR_i^*)\}_{i\in [n]}$.

\subsection{Synthetic Settings}
We test the methods in the following two types of artificial scenarios. In both scenarios, the graph is generated by the Erd\H{o}s-R\'{e}nyi model $G(n,p)$  with $n=200$ and $p=0.5$.
\subsubsection{Uniform Corruption}
We consider the following random model for generating $\bR_{ij}$:
\begin{equation}
    \bR_{ij}=\begin{cases}
    \text{Proj}(\bR_{ij}^*+\sigma \bW_{ij}),&\text{w.p. } 1-q;\\
    \tilde \bR_{ij}\sim \text{Haar}($SO(3)$),& \text{w.p. } q,
    \end{cases}
\end{equation}
where Proj denotes the projection onto $SO(3)$; $\bW_{ij}$ is a $3 \times 3$ Wigner matrix whose elements follow i.i.d.~standard normal distribution;  $\sigma\geq 0$ is a fixed noise level; $q$ is the probability that an edge is corrupted and Haar$(SO(3))$ is the  Haar probability measure on $SO(3)$. We clarify that for any $3 \times 3$ matrix $\bA$, $\text{Proj}(\bA) = \argmin_{\bR\in SO(3)} \|\bR-\bA\|_F$.

\begin{figure}[h]
    \centering
    \includegraphics[width=8cm]{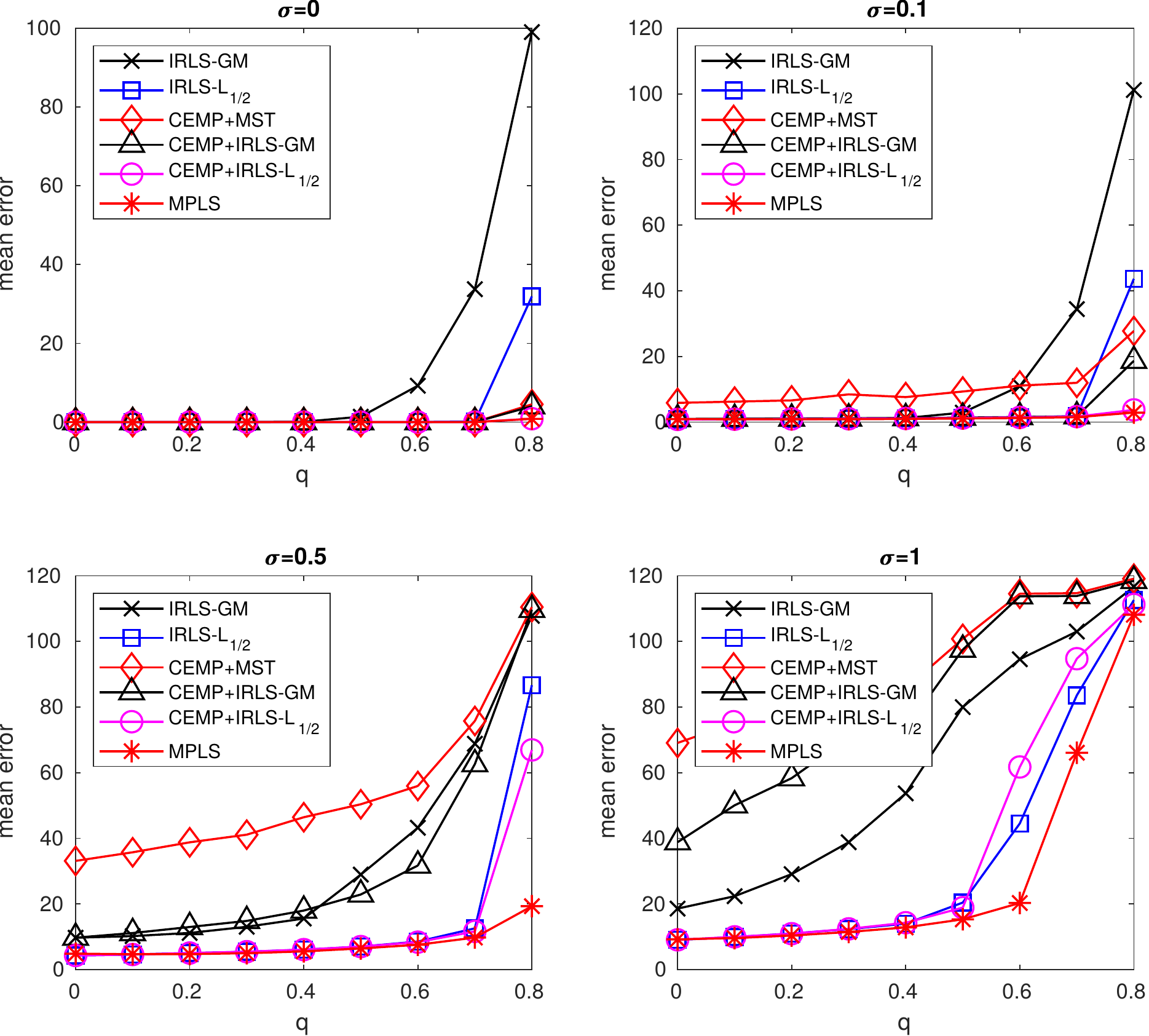}
    \caption{Performance under uniform corruption. The mean error (in degrees) is plotted against the corruption probability $q$ for 4 values of $\sigma$. }
    \label{fig:s1}
\end{figure}

We test the algorithms with four values of $\sigma:$ $0$, $0.1$, $0.5$, and $1$. We average the mean error over 10 random samples from the uniform model and report it as a function of $q$ in Figure \ref{fig:s1}.

 We note that MPLS consistently outperforms the other methods for all tested values of $q$ and $\sigma$.
 In the noiseless case, MPLS  exactly  recovers the group ratios even when $70\%$ of the edges are corrupted.
 It also nearly recovers with $80\%$ corrupted edges, where the estimation errors for IRLS-GM and IRLS-$\ell_{1/2}$ are higher than 30 degrees. MPLS is also shown to be stable under high level of noise. Since all algorithms produce poor solutions when $q=0.9$, we only show results for $0\leq q\leq 0.8$.

\begin{figure}[h]
    \centering
    \includegraphics[width=8cm]{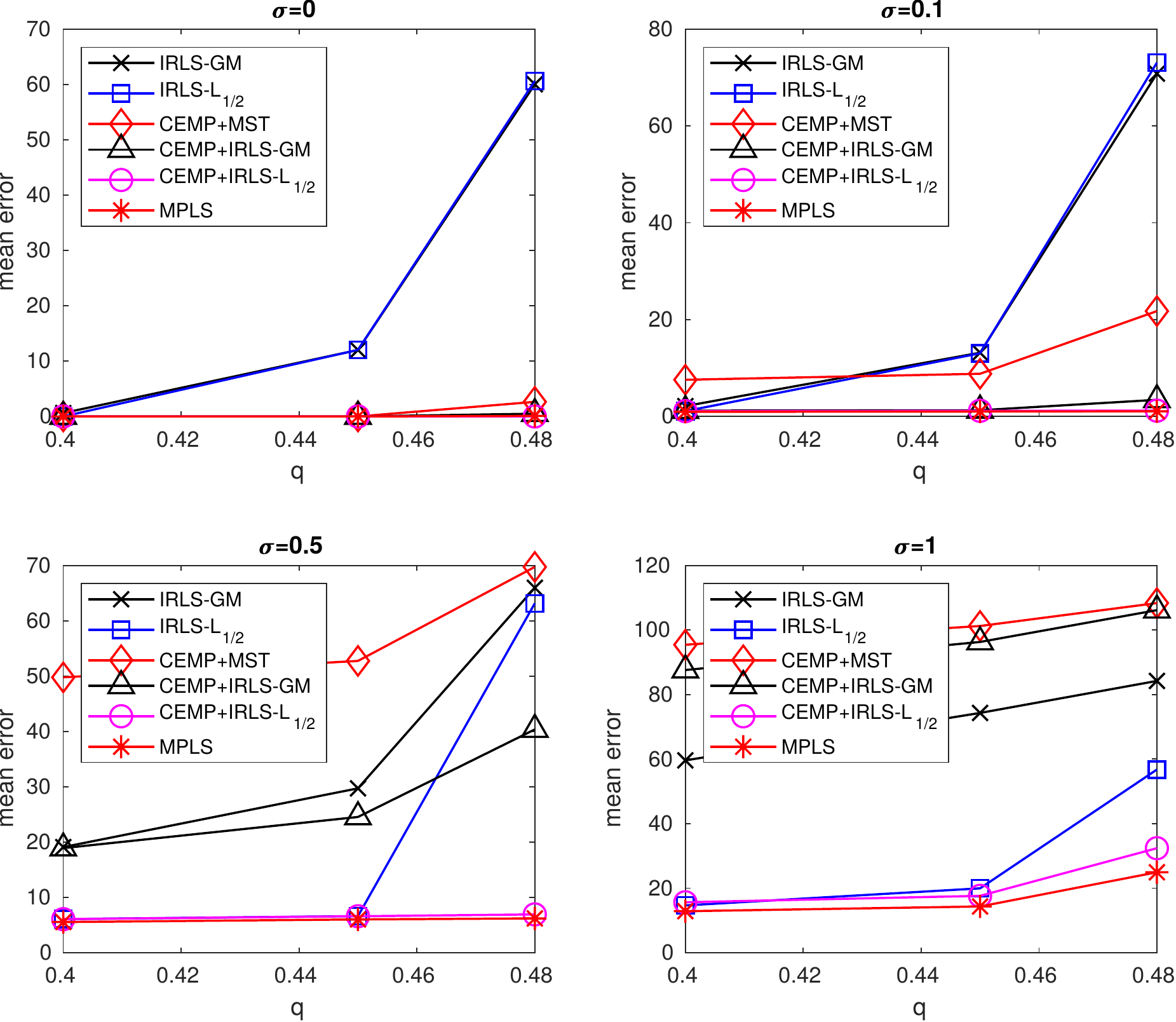}
    \caption{
    Performance under self-consistent corruption. The mean error is plotted against the corruption probability $q$ for 4 values of $\sigma$.}
    \label{fig:adv}
\end{figure}

\begin{table*}[!htbp]
\centering 
\resizebox{2\columnwidth}{!}{
\renewcommand{\arraystretch}{1.3}
\tabcolsep=0.1cm
\begin{tabular}{|l||c|c||c|c|c|c||c|c|c|c||c|c|c|c||c|c|c|c|}
\hline
Algorithms & \multicolumn{2}{c||}{}& \multicolumn{4}{c||}{IRLS-GM} &
\multicolumn{4}{c||}{IRLS-$\ell_{\frac12}$} &
 \multicolumn{4}{c||}{CEMP+MST}&
 \multicolumn{4}{c|}{MPLS} \\
\text{Dataset}& $n$ & $m$ &  {\large$\tilde{e}$} & {\large $\hat{e}$} & runtime & iter $\#$ 
& {\large$\tilde{e}$} & {\large $\hat{e}$} & runtime & iter $\#$
& {\large$\tilde{e}$} & {\large $\hat{e}$} & runtime & iter $\#$ &{\large$\tilde{e}$} & {\large $\hat{e}$} & runtime & iter $\#$\\\hline
Alamo& 564 & 71237 &
3.64 & 1.30 & 14.2 & 10+8 & 
3.67 & 1.32 & 15.5 & 10+9 & 
4.05 & 1.62 & \textbf{10.38} & 6 &
\textbf{3.44}  & \textbf{1.16}  &  20.6  & 6+8
\\\hline

Ellis Island& 223 & 17309 &
3.04 & 1.06 & 3.2 & 10+9 &
2.71 & 0.93 & 2.8 & 10+13 &
2.94 & 1.11 & \textbf{2.4} & 6 &
\textbf{2.61}  & \textbf{0.88}  & 4.0   & 6+11
\\\hline

Gendarmenmarkt& 655 & 32815&
\textbf{39.24} & \textbf{7.07} & 6.5 & 10+14 &
39.41 & 7.12 & 7.3 & 10+19 &
45.33 & 8.62 & \textbf{4.7} & 6 &
44.94 & 9.87  & 17.8  & 6+25
\\\hline

Madrid Metropolis & 315 & 14903 &
5.30 & 1.78 & 3.8 & 10+30 &
4.88 & 1.88 & 2.7 & 10+12 &
5.10 & 1.66 & \textbf{2.1} & 6 &
\textbf{4.65} & \textbf{1.26} & 5.2  & 6+23
\\\hline
Montreal N.D.& 442 & 44501 &

1.25 & 0.58 & 6.5 & 10+6 &
1.22 & 0.57 & 7.3 & 10+8 &
1.33 & 0.79 & \textbf{6.3} & 6 &
\textbf{1.04} & \textbf{0.51} & 9.3 & 6+7
\\\hline
Notre Dame & 547 & 88577 &

2.63 & 0.78 & 17.2 & 10+7 &
2.26 & 0.71 & 22.5 & 10+10 &
2.35 & 0.94 & \textbf{13.2} & 6 &
\textbf{2.06}  & \textbf{0.67}  & 31.5  & 6+8
\\\hline
NYC Library&  307 & 13814 &

2.71 & 1.37 & 2.5 & 10+14 &
2.66 & 1.30 & 2.6 & 10+15 &
3.00 & 1.41 & \textbf{1.9} & 6 &
\textbf{2.63}  & \textbf{1.24}  &  4.5   & 6+14
\\\hline
Piazza Del Popolo & 306 & 18915 &

4.10 & 2.17 & 2.8 & 10+9 &
3.99 & 2.09 & 3.1 & 10+13 &
\textbf{3.44} & \textbf{1.57} & \textbf{2.6} & 6 &
3.73 & 1.93 & 3.5 & 6+3
\\\hline
Piccadilly&  2031 & 186458 &
 5.12 & 2.02 & 153.5 & 10+16 &
 5.19 & 2.34 & 170.2 & 10+19 &
 4.66 & 1.98 & \textbf{45.8} & 6 &
 \textbf{3.93} & \textbf{1.81} & 191.9 & 6+21
 
\\\hline
Roman Forum&  989 & 41836 &

2.66 & 1.58 & 8.6 & 10+9 &
2.69 & 1.57 & 11.4 & 10+17 &
2.80 & 1.45 & \textbf{6.1} & 6 &
\textbf{2.62} & \textbf{1.37} & 8.8 & 6+8

\\\hline
Tower of London& 440 & 15918 &

3.42 & 2.52 & 2.6 & 10+8 &
3.41 & 2.50 & 2.4 & 10+12 &
\textbf{2.84} & \textbf{1.57} & \textbf{2.2} & 6 &
3.16 & 2.20 & 2.7 & 6+7

\\\hline
Union Square& 680 & 17528 &

6.77 & 3.66 & 5.0 & 10+32 &
6.77 & 3.85 & 5.6 & 10+47 &
7.47 & 3.64 & \textbf{2.5} & 6 &
\textbf{6.54} & \textbf{3.48} & 5.7 & 6+21
\\\hline
Vienna Cathedral&  770 & 87876 &

8.13 & 1.92 & 28.3 & 10+13 &
8.07 & \textbf{1.76} & 45.4 & 10+23 &
\textbf{6.91} & 2.63 & \textbf{13.1} & 6 &
7.21 & 2.83 & 42.6 & 6 +19

\\\hline
Yorkminster & 410 & 20298 &

2.60 & 1.59 & \textbf{2.4} & 10+7 &
\textbf{2.45} & 1.53 & 3.3 & 10+9 &
2.49 & \textbf{1.37} & 2.8 & 6 &
2.47 & 1.45 & 3.9 & 6+7
\\\hline

\end{tabular}}
\caption{Performance on the Photo Tourism datasets: $n$ and $m$ are the number of nodes and edges, respectively; $\tilde e$ and $\hat e$  indicate mean and median errors in degrees, respectively.; runtime is in seconds; and numbers of iterations (explained in the main text).}\label{tab:real}
\end{table*}

\subsubsection{Self-Consistent Corruption}
In order to simulate self-consistent corruption, we independently draw from Haar($SO(3)$) two classes of rotations: $\{\bR_i^*\}_{i\in [n]}$ and $\{\tilde \bR_i\}_{i\in [n]}$. We denote their corresponding relative rotations by $\bR_{ij}^*=\bR_i^*\bR_j^{*\intercal}$ and $\tilde\bR_{ij}=\tilde\bR_i\tilde\bR_j^{\intercal}$ for $ij\in E$. 
The idea is to assign to edges in $E_g$ and $E_b$ relative rotations from two different classes, so cycle-consistency occurs in both
$G([n],E_g)$ and $G([n],E_b)$. We also add noise to these relative rotations and assign them with Bernoulli model to the two classes, so one class is more significant. 
More specifically, for $ij \in E$
\begin{equation}
    \bR_{ij}=\begin{cases}
    \text{Proj}(\bR^*_{ij}+\sigma \bW_{ij}),&\text{w.p. } 1-q;\\
    \text{Proj}(\tilde\bR_{ij}+\sigma \bW_{ij}),& \text{w.p. } q,
    \end{cases}
\end{equation}
where $q$, $\sigma$, and $\bW_{ij}$ are the same as in the above uniform corruption model. We remark that an information-theoretic threshold for the exact recovery when $\sigma =0$ is $q=0.5$. That is, for $q\geq 0.5$ there is no hope of exactly recovering  $\{\bR_{i}^*\}_{i\in [n]}$.

We test the algorithms with four values of $\sigma:$ $0$, $0.1$, $0.5$, and $1$. We average the mean error over 10 random samples from the self-consistent model and report it as a function of $q$ in Figure \ref{fig:adv}.
We focus on values of  $q$ approaching the information-theoretic bound $0.5$ ($q=0.4$, $0.45$ and $0.48$). We note that MPLS consistently outperforms the other algorithm and that when $\sigma=0$ it can exactly recover the ground truth rotations when $q=0.48$.

\subsection{Real Data}
We compare the performance of the different algorithms on the Photo Tourism datasets \citep{1dsfm14}. Each of the 14 datasets consists of hundreds of 2D images of a 3D scene taken by cameras with different orientations and locations. For  each pair of images of the same scene, we use the pipeline proposed by \citet{ozyesil2015robust} to estimate the relative 3D rotations. The ground truth camera orientations are also provided.  Table \ref{tab:real}
compares the performance of IRLS-GM, IRLS-$\ell_{1/2}$, CEMP+MST and MPLS, while reporting 
mean and median errors, runtime and number of iterations. The number of iterations is the sum of the number of iterations to initialize the rotations and the number of iterations of the rest of the algorithm, where CEMP+MST only has iterations in the initialization step.

MPLS achieves the lowest mean and median error on $9$ out of $14$ datasets with runtime comparable to both IRLS, while IRLS-GM only outperforms MPLS on the Gendarmenmarkt dataset. This dataset is relatively sparse and lacks  cycle information. It contains a large amount of self-consistent corruption and none of the methods solve it reasonably well. Among the tested 4 methods, the fastest approach is CEMP+MST. 
It achieves shortest runtime on 13 out of 14 dataset. Moreover, CEMP+MST is 3 times faster than other tested methods on the largest dataset (Piccadilly). We remark that CEMP+MST is able to achieve comparable results to common IRLS on most datasets, and has superior performance on 2 datasets, which have some perfectly estimated edges.
In summary, for most of the datasets, MPLS provides the highest accuracy and CEMP+MST obtains the fastest runtime. 
\section{Conclusion}
\label{sec:conclusion}

We proposed a framework for solving group synchronization under high corruption and noise. 
This general framework requires a successful solution of the weighted least squares problem, which depends on the group. For $SO(3)$, we explained how a well-known solution integrates well with our framework. We demonstrated state-of-the-art performance of our framework for $SO(3)$ synchronization. We have motivated our method as an alternative to IRLS and explained how it may overcome the limitations of IRLS when applied to group synchronization.

There are many directions to expand 
our work. One can carefully adapt and implement our proposed framework to other groups that occur in practice. 
One may develop certain theoretical guarantees for convergence and exact recovery of MPLS.

\section*{Acknowledgement}
This work was supported by NSF award DMS-18-21266. We thank Tyler Maunu for his valuable comments
on an earlier version of this manuscript.



\bibliography{cemp}

\begin{thebibliography}{37}
\providecommand{\natexlab}[1]{#1}
\providecommand{\url}[1]{\texttt{#1}}
\expandafter\ifx\csname urlstyle\endcsname\relax
  \providecommand{\doi}[1]{doi: #1}\else
  \providecommand{\doi}{doi: \begingroup \urlstyle{rm}\Url}\fi

\bibitem[Abbe(2017)]{Z2abbe}
Abbe, E.
\newblock Community detection and stochastic block models: recent developments.
\newblock \emph{The Journal of Machine Learning Research}, 18\penalty0
  (1):\penalty0 6446--6531, 2017.

\bibitem[Abbe et~al.(2014)Abbe, Bandeira, Bracher, and Singer]{Z2Afonso2}
Abbe, E., Bandeira, A.~S., Bracher, A., and Singer, A.
\newblock Decoding binary node labels from censored edge measurements: Phase
  transition and efficient recovery.
\newblock \emph{{IEEE} Trans. Network Science and Engineering}, 1\penalty0
  (1):\penalty0 10--22, 2014.

\bibitem[Arrigoni et~al.(2016)Arrigoni, Rossi, and Fusiello]{SE3_sync}
Arrigoni, F., Rossi, B., and Fusiello, A.
\newblock Spectral synchronization of multiple views in {SE(3)}.
\newblock \emph{{SIAM} J. Imaging Sciences}, 9\penalty0 (4):\penalty0
  1963--1990, 2016.

\bibitem[Arrigoni et~al.(2018)Arrigoni, Rossi, Fragneto, and
  Fusiello]{SE3_RPCA}
Arrigoni, F., Rossi, B., Fragneto, P., and Fusiello, A.
\newblock Robust synchronization in {SO(3)} and {SE(3)} via low-rank and sparse
  matrix decomposition.
\newblock \emph{Comput. Vis. Image Underst.}, 174:\penalty0 95--113, 2018.

\bibitem[Bandeira(2018)]{Z2Afonso}
Bandeira, A.~S.
\newblock Random laplacian matrices and convex relaxations.
\newblock \emph{Foundations of Computational Mathematics}, 18\penalty0
  (2):\penalty0 345--379, 2018.
\newblock \doi{10.1007/s10208-016-9341-9}.

\bibitem[Bandeira et~al.(2017)Bandeira, Boumal, and Singer]{rotationNP}
Bandeira, A.~S., Boumal, N., and Singer, A.
\newblock Tightness of the maximum likelihood semidefinite relaxation for
  angular synchronization.
\newblock \emph{Mathematical Programming}, 163\penalty0 (1-2):\penalty0
  145--167, 2017.

\bibitem[Birdal et~al.(2018)Birdal, Simsekli, Eken, and Ilic]{SE3_MCMC}
Birdal, T., Simsekli, U., Eken, M.~O., and Ilic, S.
\newblock Bayesian pose graph optimization via bingham distributions and
  tempered geodesic {MCMC}.
\newblock In \emph{Advances in Neural Information Processing Systems 31: Annual
  Conference on Neural Information Processing Systems (NeurIPS)}, pp.\
  306--317, 2018.

\bibitem[Briales \& Jim{\'{e}}nez(2017)Briales and
  Jim{\'{e}}nez]{SE3_SDP_jesus}
Briales, J. and Jim{\'{e}}nez, J.~G.
\newblock Cartan-sync: Fast and global se(d)-synchronization.
\newblock \emph{{IEEE} Robotics Autom. Lett.}, 2\penalty0 (4):\penalty0
  2127--2134, 2017.

\bibitem[Chatterjee \& Govindu(2013)Chatterjee and
  Govindu]{ChatterjeeG13_rotation}
Chatterjee, A. and Govindu, V.~M.
\newblock Efficient and robust large-scale rotation averaging.
\newblock In \emph{{IEEE} International Conference on Computer Vision, {ICCV}
  2013, Sydney, Australia, December 1-8, 2013}, pp.\  521--528, 2013.

\bibitem[Chatterjee \& Govindu(2018)Chatterjee and Govindu]{L12}
Chatterjee, A. and Govindu, V.~M.
\newblock Robust relative rotation averaging.
\newblock \emph{{IEEE} Trans. Pattern Anal. Mach. Intell.}, 40\penalty0
  (4):\penalty0 958--972, 2018.
\newblock \doi{10.1109/TPAMI.2017.2693984}.

\bibitem[Chen et~al.(2014)Chen, Guibas, and Huang]{chen_partial}
Chen, Y., Guibas, L.~J., and Huang, Q.
\newblock Near-optimal joint object matching via convex relaxation.
\newblock In \emph{Proceedings of the 31th International Conference on Machine
  Learning, {ICML} 2014, Beijing, China, 21-26 June 2014}, pp.\  100--108,
  2014.

\bibitem[Donoho et~al.(2009)Donoho, Maleki, and Montanari]{AMP_Donoho}
Donoho, D.~L., Maleki, A., and Montanari, A.
\newblock Message-passing algorithms for compressed sensing.
\newblock \emph{Proceedings of the National Academy of Sciences}, 106\penalty0
  (45):\penalty0 18914--18919, 2009.
\newblock ISSN 0027-8424.
\newblock \doi{10.1073/pnas.0909892106}.

\bibitem[Eriksson et~al.(2018)Eriksson, Olsson, Kahl, and
  Chin]{StrongDuality18}
Eriksson, A.~P., Olsson, C., Kahl, F., and Chin, T.
\newblock Rotation averaging and strong duality.
\newblock In \emph{2018 {IEEE} Conference on Computer Vision and Pattern
  Recognition, {CVPR} 2018, Salt Lake City, UT, USA, June 18-22}, pp.\
  127--135. {IEEE} Computer Society, 2018.

\bibitem[Gao \& Zhao(2019)Gao and Zhao]{ICMLphase}
Gao, T. and Zhao, Z.
\newblock Multi-frequency phase synchronization.
\newblock In \emph{Proceedings of the 36th International Conference on Machine
  Learning, {ICML} 2019, 9-15 June 2019, Long Beach, California, {USA}}, pp.\
  2132--2141, 2019.

\bibitem[Govindu(2004)]{Govindu04_Lie}
Govindu, V.~M.
\newblock Lie-algebraic averaging for globally consistent motion estimation.
\newblock In \emph{2004 {IEEE} Computer Society Conference on Computer Vision
  and Pattern Recognition {(CVPR} 2004), 27 June - 2 July 2004, Washington, DC,
  {USA}}, pp.\  684--691, 2004.

\bibitem[Hand et~al.(2018)Hand, Lee, and Voroninski]{HandLV15}
Hand, P., Lee, C., and Voroninski, V.
\newblock Shapefit: Exact location recovery from corrupted pairwise directions.
\newblock \emph{Communications on Pure and Applied Mathematics}, 71\penalty0
  (1):\penalty0 3--50, 2018.

\bibitem[Hartley et~al.(2011)Hartley, Aftab, and Trumpf]{HartleyAT11_rotation}
Hartley, R.~I., Aftab, K., and Trumpf, J.
\newblock {L1} rotation averaging using the weiszfeld algorithm.
\newblock In \emph{The 24th {IEEE} Conference on Computer Vision and Pattern
  Recognition, {CVPR}}, pp.\  3041--3048, 2011.

\bibitem[Hartley et~al.(2013)Hartley, Trumpf, Dai, and Li]{RotationAveraging13}
Hartley, R.~I., Trumpf, J., Dai, Y., and Li, H.
\newblock Rotation averaging.
\newblock \emph{Int. J. Comput. Vis.}, 103\penalty0 (3):\penalty0 267--305,
  2013.
\newblock \doi{10.1007/s11263-012-0601-0}.

\bibitem[Huang \& Guibas(2013)Huang and Guibas]{Huang13}
Huang, Q. and Guibas, L.~J.
\newblock Consistent shape maps via semidefinite programming.
\newblock \emph{Comput. Graph. Forum}, 32\penalty0 (5):\penalty0 177--186,
  2013.

\bibitem[Huang et~al.(2017)Huang, Liang, Bajaj, and Huang]{truncatedLS}
Huang, X., Liang, Z., Bajaj, C., and Huang, Q.
\newblock Translation synchronization via truncated least squares.
\newblock In \emph{Advances in Neural Information Processing Systems 30: Annual
  Conference on Neural Information Processing Systems}, pp.\  1459--1468, 2017.

\bibitem[Huroyan(2018)]{PPM_vahan}
Huroyan, V.
\newblock \emph{Mathematical Formulations, Algorithm and Theory for Big Data
  Problems}.
\newblock PhD thesis, University of Minnesota, 2018.

\bibitem[Lerman \& Shi(2019)Lerman and Shi]{cemp}
Lerman, G. and Shi, Y.
\newblock Robust group synchronization via cycle-edge message passing.
\newblock \emph{arXiv preprint arXiv:1912.11347}, 2019.

\bibitem[Lerman et~al.(2018)Lerman, Shi, and Zhang]{LUDrecovery}
Lerman, G., Shi, Y., and Zhang, T.
\newblock Exact camera location recovery by least unsquared deviations.
\newblock \emph{{SIAM} J. Imaging Sciences}, 11\penalty0 (4):\penalty0
  2692--2721, 2018.
\newblock \doi{10.1137/17M115061X}.

\bibitem[Maunu \& Lerman(2020)Maunu and Lerman]{SO2ML}
Maunu, T. and Lerman, G.
\newblock A provably robust multiple rotation averaging scheme for {SO(2)}.
\newblock \emph{arXiv preprint arXiv:2002.05299}, 2020.

\bibitem[Ozyesil \& Singer(2015)Ozyesil and Singer]{ozyesil2015robust}
Ozyesil, O. and Singer, A.
\newblock Robust camera location estimation by convex programming.
\newblock In \emph{Proceedings of the IEEE Conference on Computer Vision and
  Pattern Recognition}, pp.\  2674--2683, 2015.

\bibitem[Pachauri et~al.(2013)Pachauri, Kondor, and Singh]{deepti}
Pachauri, D., Kondor, R., and Singh, V.
\newblock Solving the multi-way matching problem by permutation
  synchronization.
\newblock In Burges, C. J.~C., Bottou, L., Welling, M., Ghahramani, Z., and
  Weinberger, K.~Q. (eds.), \emph{Advances in Neural Information Processing
  Systems 26}, pp.\  1860--1868. Curran Associates, Inc., 2013.

\bibitem[{Perry} et~al.(2018){Perry}, {Wein}, {Bandeira}, and
  {Moitra}]{AMP_compact}
{Perry}, A., {Wein}, A.~S., {Bandeira}, A.~S., and {Moitra}, A.
\newblock {Message-passing algorithms for synchronization problems over compact
  groups}.
\newblock \emph{Communications on Pure and Applied Mathematics}, 2018.

\bibitem[Purkait et~al.(2019)Purkait, Chin, and Reid]{NeuroRA}
Purkait, P., Chin, T., and Reid, I.~D.
\newblock Neurora: Neural robust rotation averaging.
\newblock \emph{arXiv preprint arXiv:1912.04485}, 2019.

\bibitem[Rosen et~al.(2019)Rosen, Carlone, Bandeira, and Leonard]{SE3_Rosen}
Rosen, D.~M., Carlone, L., Bandeira, A.~S., and Leonard, J.~J.
\newblock Se-sync: {A} certifiably correct algorithm for synchronization over
  the special euclidean group.
\newblock \emph{I. J. Robotics Res.}, 38\penalty0 (2-3), 2019.

\bibitem[Shen et~al.(2016)Shen, Zhu, Fang, Zhang, and Quan]{shen2016}
Shen, T., Zhu, S., Fang, T., Zhang, R., and Quan, L.
\newblock Graph-based consistent matching for structure-from-motion.
\newblock In \emph{European Conference on Computer Vision}, pp.\  139--155.
  Springer, 2016.

\bibitem[Shi \& Lerman(2018)Shi and Lerman]{AAB}
Shi, Y. and Lerman, G.
\newblock Estimation of camera locations in highly corrupted scenarios: All
  about that base, no shape trouble.
\newblock In \emph{2018 {IEEE} Conference on Computer Vision and Pattern
  Recognition, {CVPR}}, pp.\  2868--2876, 2018.

\bibitem[Shi et~al.(2020)Shi, Li, and Lerman]{robust_multi_object2020}
Shi, Y., Li, S., and Lerman, G.
\newblock Robust multi-object matching via iterative reweighting of the graph
  connection laplacian, 2020.

\bibitem[Tron et~al.(2008)Tron, Vidal, and Terzis]{consensusSO3}
Tron, R., Vidal, R., and Terzis, A.
\newblock Distributed pose averaging in camera networks via consensus on
  {SE(3)}.
\newblock In \emph{2008 Second {ACM/IEEE} International Conference on
  Distributed Smart Cameras, Stanford, CA, USA, September 7-11, 2008}, pp.\
  1--10, 2008.

\bibitem[Wang \& Singer(2013)Wang and Singer]{wang2013exact}
Wang, L. and Singer, A.
\newblock Exact and stable recovery of rotations for robust synchronization.
\newblock \emph{Information and Inference}, 2013.

\bibitem[Wilson \& Snavely(2014)Wilson and Snavely]{1dsfm14}
Wilson, K. and Snavely, N.
\newblock Robust global translations with 1dsfm.
\newblock In \emph{Computer Vision - {ECCV} 2014 - 13th European Conference,
  Zurich, Switzerland, September 6-12, 2014, Proceedings, Part {III}}, pp.\
  61--75, 2014.

\bibitem[Yedidia et~al.(2003)Yedidia, Freeman, and Weiss]{BP}
Yedidia, J.~S., Freeman, W.~T., and Weiss, Y.
\newblock Understanding belief propagation and its generalizations.
\newblock \emph{Exploring artificial intelligence in the new millennium},
  8:\penalty0 236--239, 2003.

\bibitem[Zach et~al.(2010)Zach, Klopschitz, and Pollefeys]{Zach2010}
Zach, C., Klopschitz, M., and Pollefeys, M.
\newblock Disambiguating visual relations using loop constraints.
\newblock In \emph{The Twenty-Third {IEEE} Conference on Computer Vision and
  Pattern Recognition, {CVPR}}, pp.\  1426--1433, 2010.

\end{thebibliography}
\bibliographystyle{icml2020}

\end{document}